\documentclass[10pt,twocolumn,letterpaper,pagenumbers]{article}

\usepackage[pagenumbers]{cvpr} 

\usepackage{multirow}
\usepackage[T1]{fontenc}

\definecolor{cvprblue}{rgb}{0.21,0.49,0.74}
\usepackage[pagebackref,breaklinks,colorlinks,allcolors=cvprblue]{hyperref}

\def\paperID{****} 
\def\confName{****}
\def\confYear{****}

\title{Text-promptable Object Counting via Quantity Awareness Enhancement}

\author{Miaojing Shi$^{1,4}$,
Xiaowen Zhang$^1$,
Zijie Yue$^{1}$\thanks{Corresponding author: zijie@tongji.edu.cn} ,
Yong Luo$^2$,
Cairong Zhao$^3$,
Li Li$^{1}$\\
$^1$College of Electronic and Information Engineering, Tongji University.\\
$^2$College of Computer Science, Wuhan University.\\
$^3$College of Computer Science, Tongji University.\\
$^4$State Key Laboratory of Autonomous Intelligent Unmanned Systems.\\
}

\begin{document}
\maketitle
\begin{abstract}
Recent advances in large vision-language models (VLMs) have shown remarkable progress in solving the  text-promptable object counting problem. Representative methods typically specify text prompts with object category information in images. This however is insufficient for training the model to accurately distinguish the number of objects in the counting task. To this end, we propose QUANet, which introduces novel quantity-oriented text prompts with a vision-text quantity alignment loss to enhance the model's quantity awareness. Moreover, we propose a dual-stream adaptive counting decoder consisting of a Transformer stream, a CNN stream, and a number of Transformer-to-CNN enhancement adapters (T2C-adapters) for density map prediction. The T2C-adapters facilitate the effective knowledge communication and aggregation between the Transformer and CNN streams. A cross-stream quantity ranking loss is proposed in the end to optimize the ranking orders of predictions from the two streams. Extensive experiments on standard benchmarks such as FSC-147, CARPK, PUCPR+, and ShanghaiTech demonstrate our model’s strong generalizability for zero-shot class-agnostic counting. 
Code is available at \url{https://github.com/viscom-tongji/QUANet}
\end{abstract}    
\section{Introduction}
\label{sec:intro}
The object counting task was initially developed to estimate the quantity of specific targets within images or videos. Conventional methods achieve it by training deep neural networks to estimate object density maps on vast amounts of labeled data from predefined categories, such as crowds~\citep{liang2022end}, vehicles~\citep{mundhenk2016large}, and cells~\citep{xie2018microscopy}. Despite their achievements, these methods lack the flexibility to generalize to previously unseen categories. Recent methods introduce class-agnostic counting~\citep{lu2019class,ranjan2022exemplar}, allowing for the counting of arbitrary categories by using exemplars as category indicators. By annotating a few image patches as exemplars and computing the similarities between exemplars and image regions, these methods have demonstrated promising accuracy as well as good generalizability. However, they still depend on the manual annotation for objects of interest, which is often inconvenient in real-world applications.

\begin{figure}
  \centering
  \includegraphics[width=0.5\textwidth]{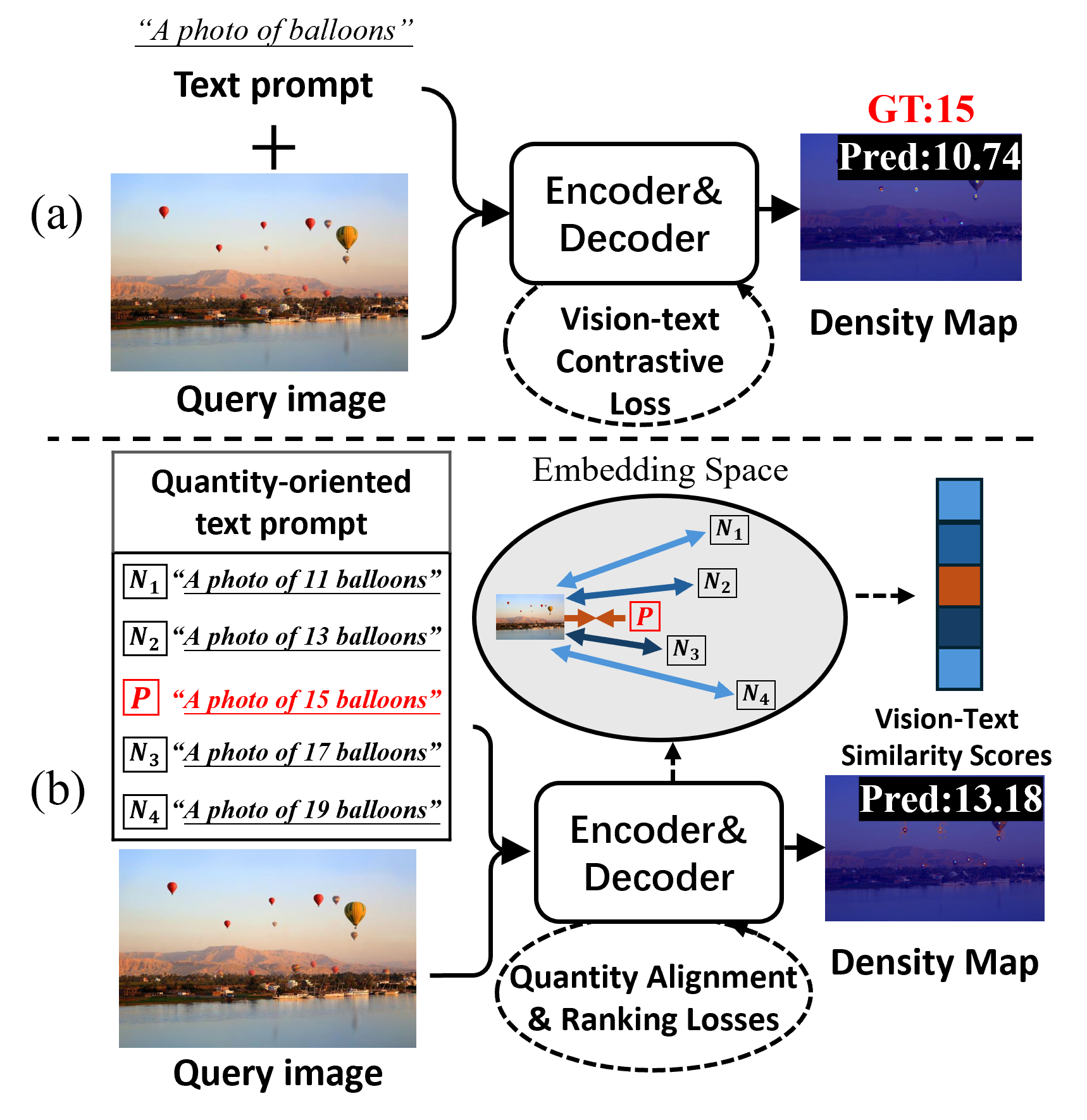}
   \caption{
   (a) Previous text-promptable object counting methods use simple text prompts and conventional vision-text contrastive loss. (b) Our QUANet introduces novel quantity-oriented text prompts with quantity alignment and ranking losses to enhance the model's quantity awareness.}
   \label{fig:openfig}
\end{figure}

Recently, pre-trained large vision-language models (VLMs) ~\citep{liu2024visual,li2023blip} have proven effective in numerous vision tasks. 
Inspired by their success,
there has been a surge of interest in solving the class-agnostic counting problem based on VLMs, \ie leveraging text prompts instead of exemplars to specify the target category of interest~\citep{kang2024vlcounter,jiang2023clip,AminiNaieni23,10740342}. 
The text prompts are generated either manually~\citep{kang2024vlcounter,AminiNaieni23} or automatically~\citep{doubinsky2024semantic}. Both texts and images are encoded via the VLM to obtain their respective embeddings and then interacted for object count prediction.
In this way, they achieve zero-shot object counting without visual exemplars. The performance of this text-promptable object counting highly relies on the quality of text prompts. 
Existing works use simple prompts or elaborate category-related details, for instance, using the color and position descriptions to specify target object categories ~\citep{dai2024referring}.
While this is important for object recognition, it is however insufficient for the model to distinguish the number of objects in the counting task. 
\cite{paiss2023teaching} instead alters the numerical information in text prompts to predefined numbers (up to 10), so as to leverage them as positive/negative vision-text pairs in the contrastive learning~\citep{paiss2023teaching,liang2023crowdclip}.     
This reassures the model's emphasis on the quantity side, yet is not enough. It only allows the model to coarsely know that these numerical values in texts correspond to different numbers of objects in images. Knowing how much is the difference between numbers, as the key to exact object counting, was however untouched in previous works. 
To achieve it, we propose novel quantity-oriented text prompts to dynamically manipulate the numerical information, and through a series of quantity alignment and ranking losses, we optimize the model to comprehensively understand object quantities in images  
(Fig.~\ref{fig:openfig}).

Next, many object counting methods have chosen the Transformer-based encoders for robust feature encoding, yet when it comes to the object density estimation stage, CNN-based decoders remain the only  choice~\citep{jiang2023clip,AminiNaieni23,liu2022countr}. CNNs are effective in extracting and processing local features, which can be especially beneficial to density estimation-based object counting. The Transformer (\eg ViT~\citep{dosovitskiy2020image}), in spite of its global feature modeling capability, has not been successfully adopted in object density estimation~\citep{liu2022countr,kang2024vlcounter}.   This can be attributed to the complication of the density estimation task, where the complementary characteristics between CNN and Transformer are imperfectly reflected in individual density maps. To cope with it, we empower a novel dual-stream adaptive counting decoder, where 
several Transformer-to-CNN enhancement adapters (T2C-adapters) and a gating network are specifically developed to transform the complementary nature between CNN and Transformer into realistic benefit for object counting.

In this paper, we introduce the \textbf{QUANet}, a \textbf{QU}antity-\textbf{A}ware neural \textbf{Net}work for text-promptable object counting. 
During network training, given an image, we craft the {quantity-oriented text prompts} that consist of 
the factual text prompt specifying the exact object count and counterfactual text prompts specifying incorrect counts according to dynamically assigned intervals. 
Next, we pass these prompts and the image into the VLM's encoders to obtain textual and visual embeddings, respectively. 
A {dual-stream adaptive counting decoder} is proposed to decode the density map, which consists of a Transformer (ViT) stream and a CNN stream to capture complementary features. We design several T2C-adapters and a gating net to enable effective knowledge communication and aggregation among these features, ultimately improving the counting performance. 
In the end, we propose the vision-text quantity alignment loss to enforce similarities between vision-text pairs to conform to specific quantity order given in text prompts; this is to inject the awareness of quantity differences into the visual counter. The idea is further reinforced via a new cross-stream quantity ranking loss, which enforces the patch-level counting consistency and ranking across the two streams of the decoder. During testing, {quantity-oriented text prompts} are unavailable; instead, a conventional category-specific text prompt is used.  

Extensive experiments on FSC-147~\citep{ranjan2021learning}, CARPK~\citep{hsieh2017drone}, PUCPR+~\citep{hsieh2017drone}, and ShanghaiTech~\citep{zhang2016single} datasets demonstrate that our method outperforms state-of-the-art text-promptable object counting methods by a large margin.

\section{Related work}
\label{sec:Related work}
\subsection{Class-specific Object Counting}

Class-specific object counting focuses on counting specific categories from given images. These categories mainly include crowds~\citep{ranasinghe2024crowddiff,guo2024regressor,peng2024single,wu2023boosting,du2023domain,liu2022discovering,shi2019revisiting}, cars \citep{hsieh2017drone,mundhenk2016large}, cells~\citep{xie2018microscopy}, and animals~\citep{arteta2016counting}. The solutions can be divided into detection-based~\citep{liu2023point,liang2022end,liu2019point} and regression-based~\citep{ranasinghe2024crowddiff,guo2024regressor,peng2024single,wu2023boosting,li2023calibrating} methods. Detection-based methods employ object detection models to generate object bounding boxes of target categories, where the total number of bounding boxes corresponds to the object count. 
{However, these solutions easily struggle in scenarios with small, overlapping, or densely packed objects. As a result, regression-based methods, which primarily rely on the density estimation for object counting, have become the mainstream. }
The target count is obtained by integrating the pixel values over the estimated density map.

\subsection{Class-agnostic Object Counting}
Class-agnostic counting aims to count objects of arbitrary categories within images without requiring specific training for each category. They normally utilize additional visual exemplars to indicate the target category and regard the number of local patches with high similarities to exemplars as the final count~\citep{lu2019class,shi2022represent,liu2022countr,ranjan2021learning,djukic2023low,pelhan2024dave,chen2025single,yang2025pbecount}. For example, GMN~\citep{lu2019class} was the first to formulate the class-agnostic counting task as a feature matching problem between the image and the exemplar. CounTR~\citep{liu2022countr} utilizes a feature interaction module to fuse the image features and exemplar features for density map prediction. However, these methods require manual selection or annotation of visual exemplars, which can be inconvenient. 
To address this issue, exemplar-free methods such as RepRPN-Counter~\citep{ranjan2022exemplar} have been proposed to predict the count of the most frequently appearing objects in the image. Nevertheless, these methods typically focus on counting only the salient objects in images rather than on specific categories of interests.

\subsection{Text-promptable Object Counting}

Text-promptable object counting~\citep{Xu_2023_CVPR,kang2024vlcounter,jiang2023clip,AminiNaieni23,zhu2024zero,10740342} has emerged as a popular class-agnostic counting method in recent years. It uses text prompts as substitutes of the visual exemplars to guide models in counting specified object categories within images, enhancing flexibility and reducing reliance on manual annotations. For example, CLIP-count~\citep{jiang2023clip} first utilizes the pre-trained CLIP~\citep{radford2021learning} model to extract visual and textual embeddings, which are then fused via a hierarchical text-patch interaction module for density map estimation. VLCounter ~\citep{kang2024vlcounter} employs a semantic-conditioned prompt tuning mechanism to inject category information from the prompt into the visual embeddings, and uses a CNN-based decoder to predict object counts. CounTX~\citep{AminiNaieni23} enhances the open-set counting capability by using Open-CLIP~\citep{cherti2023reproducible} as the backbone network. ~\cite{dai2024referring} and ~\cite{wang2025exploring} have additionally annotated local object attributes in images to augment category-related prompts; the object quantity information, though important, is not emphasized. In contrast, ~\cite{paiss2023teaching} changes the numerical values in text prompts indiscriminately to predefined numbers (1-10) to create negative samples for conventional contrastive learning. This way gives the model certain sense of quantity, which however is very coarse, as they apply a uniform treatment to those  negative samples with different numerical values. Also, ~\cite{paiss2023teaching} can only count objects up to 10. We believe the core of quantity awareness is knowing how much is the difference among numbers in images and we have thereby introduced a number of new components to realize it.  

\begin{figure*}[t]
  \centering
  \includegraphics[width=1.0\textwidth]{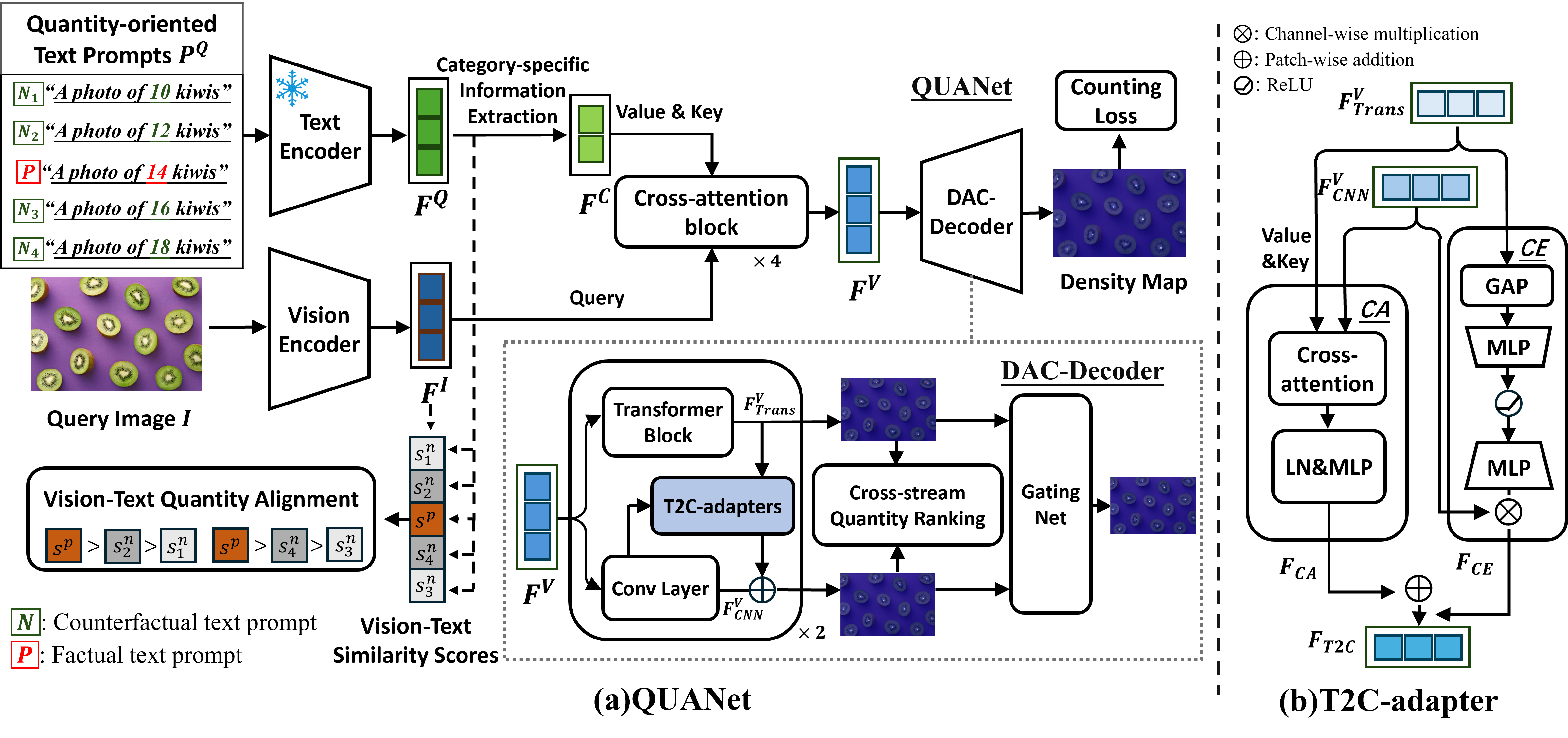}
   \caption{(a) Overall training architecture of our QUANet. Given a query image, we first craft the quantity-oriented text prompts to specify the numbers of objects for the target category. We design the factual text prompt specifying the exact object count and counterfactual text prompts specifying incorrect counts. These prompts and the image are then fed into the VLM’s encoders to obtain textual and visual embeddings. We develop a novel vision-text quantity alignment loss to inject quantity awareness into the vision encoder. Next, we extract category-specific information from the textual embedding and integrate it into the visual embedding. A DAC-decoder is then proposed to predict the density map using a Transformer stream, a CNN stream, and a number of T2C-adapters. A cross-stream quantity ranking loss is proposed in the end to optimize the ranking orders of predictions from the two streams. (b) The structure of T2C-adapter. It leverages a cross-attention block and a channel-excitation block to transfer the knowledge from the Transformer stream into the CNN stream.} \vspace{-10pt}
   \label{fig:overview}
\end{figure*}

\section{Method}
\subsection{Overview}
Given an image $I \in R^{H\times W \times 3 }$ and a category label $C$, we aim to output the density map $D^{*}\in R^{H\times W \times 1 }$ corresponding to $C$. The training framework of our proposed QUANet is shown in Fig.~\ref{fig:overview}(a).
First, we generate the quantity-oriented text prompts $P^{Q}$ to describe the numbers of objects of the target category in $I$. 
We design the factual prompt specifying the exact object count and counterfactual prompts with incorrect counts. Combining these prompts with the image $I$, we can obtain positive (matched) and negative (unmatched) vision-text pairs, respectively. Textual embeddings $F^{Q}$ and visual embeddings $F^{I}$ are extracted via the VLM's encoders. 
At the decoding stage, we design a Dual-stream Adaptive Counting decoder (DAC-decoder) (Sec.~\ref{GIT decoder}) for density map $D^{*}$ prediction, which processes complementary features from Transformer (ViT) and CNN streams. Several layer-wise Transformer-to-CNN enhancement adapters (T2C-adapters) and a gating net by the end are proposed to enable the interaction and aggregation between the two streams. Finally, we introduce a novel vision-text quantity alignment loss (Sec.~\ref{Network optimization}) to inject quantity awareness into the vision encoder; and a cross-stream quantity ranking loss (Sec.~\ref{Network optimization}) to reinforce the quantity awareness and consistency across the two streams of the decoder.

\subsection{Quantity-oriented text prompts}\label{Prompts generate}
Previous text-promptable object counting methods typically use simple prompts (\eg a photo of [class]) to train the models, which is insufficient to distinguish the number of objects. 
~\cite{paiss2023teaching} augments the prompts by changing the numerical values inside, but they do it merely to fill in the negative terms in the conventional contrastive learning, without fostering a deeper understanding of numerical concepts. We design the quantity-oriented text prompts $P^{Q}$ to inject the model sophisticated quantity awareness. 

The template of $P^{Q}$ is ``a photo of [num] [class]'', where the [class] token is replaced with the target category name, while the [num] token is replaced with the ground truth count $a^p$ of the target object in  $I$ to construct a \emph{factual text prompt}. Furthermore, we also replace the [num] token with alternative counts $ \left \{ a^p\pm k\Delta | k \in [1,\frac{N}{2}] \right \}$
to generate a series of \emph{counterfactual text prompts}, where $\Delta $ is a dynamical value interval based on $a^p$ (see Sec~\ref{Experiments details}).
{For instance, as shown in Fig.~\ref{fig:overview}(a), given an image with 14 kiwis, the factual text prompt is formulated as ``a photo of 14 kiwis'', while the counterfactual  text prompts are created by altering the count to ``a photo of 10/12/16/18 kiwis'', respectively.}
We regard $I$ combined with the factual text prompt as a positive vision-text pair, while $I$ combined with the counterfactual text prompts as negative vision-text pairs. 
Later in Sec.~\ref{Network optimization}, we introduce the vision-text quantity alignment loss and cross-stream quantity ranking loss to optimize the vision-text similarities to enhance the model's quantity awareness.

Besides, to guide the model to predict the density map for the specified object category, we also need to extract the category-related information from $F^{Q}$. We do this by removing the word embedding of the [num] token from $F^{Q}$ to obtain $F^{C}$. Following CounTR~\citep{liu2022countr}, 
four cross-attention blocks are leveraged to facilitate interactions between $F^{I}$ and $F^{C}$ to generate the category-prompted visual feature $F^V$ for subsequent density map prediction (Sec.~\ref{GIT decoder}).

\subsection{Dual-stream Adaptive Counting decoder}
\label{GIT decoder}
CNN-based decoders are particularly favored in the density-estimation based counting task~\citep{jiang2023clip,AminiNaieni23,liu2022countr}, while Transformer-based decoders, though effective in capturing global semantic information, have not been successfully transferred into practical usage. Unlike detection-based methods, density estimation-based methods do not rely on strong semantic information. The complementary characteristics between CNN and Transformer, though exist in general, present clear variations in individual density maps. To cope with this situation, 
as shown in Fig.\ref{fig:overview}, we propose a DAC-decoder which devises a Transformer stream in parallel to the CNN stream,  
while several layer-wise T2C-adapters are introduced to effectively transfer the knowledge from the Transformer stream into the CNN stream on the patch level.  
The outputs of two streams are projected into density maps $D^*_{CNN}$ and $D^*_{Trans}$ via $1\times1$
convolutional layers, respectively. A gating net is developed to further fuse $D^*_{CNN}$ and $D^*_{Trans}$ on the image level for final density map prediction. 
Below we specify the details of T2C-adapters and the gating net.

\noindent\textbf{T2C-adapter.} The structure of T2C-adapter is shown in Fig.\ref{fig:overview}(b). It has a Cross-Attention block (CA) and a Channel-Excitation block (CE). {The former primarily facilitates interaction between the two streams in the DAC-decoder along the spatial dimension, while the latter is proposed to further enhance the feature representation by recalibrating channel-wise feature responses.} Specifically, the CA block is comprised of a cross-attention layer, a layer normalization layer, and a MLP layer. For the cross-attention layer, we treat the global feature extracted from the Transformer stream as the key and value, while the local features from the CNN stream as the query.
This process is written as: $F_{CA}=CA(F^{V}_{CNN},F^{V}_{Trans})$, where $F^{V}_{CNN}$ and $F^{V}_{Trans}$ represent the feature $F^{V}$ being processed by the convolutional layer and the Transformer block respectively in the two streams. 

The CE block, inspired by the Squeeze-and-Excitation (SE) block~\citep{hu2018squeeze}, is designed to recalibrate the importance of local features along the channel dimension. Unlike the SE block, which calculates channel-wise attention weights from the input feature map and applies them back to the input itself, our CE block utilizes the feature $F^V_{Trans}$ from the Transformer stream and projects it via global average pooling (GAP) into an attention vector to re-weight the CNN feature $F^V_{CNN}$ channel-wisely. This process modulates $F^V_{CNN}$ to emphasize those channels with higher relevance to $F^V_{Trans}$ while attenuating less relevant ones. It is expressed as: $F_{CE}=CE(F^{V}_{CNN},F^{V}_{Trans})$. The final output of the T2C-adapter is $F_{T2C} = F_{CE}+F_{CA}$, which is added to $F^V_{CNN}$. 

Note that we adapt the knowledge extracted from the Transformer stream into that of the CNN stream in a uni-directional way. We find that 
the opposite way would break the contextual cues originally stored in the Transformer stream. 
{The experimental analysis in Sec.~\ref{Ablation studies} verifies this.}

\noindent\textbf{Gating net.} We develop a gating net to aggregate the outputs of the CNN and Transformer streams ($D^{*}_{CNN}$ and $D^{*}_{Trans}$). It learns weights for $D^{*}_{CNN}$ and $D^{*}_{Trans}$ from the original category-prompted visual feature $F^{V}$. The two density maps are then fused via the weighted sum to obtain the final prediction: $D^{*}$ =  $w_1 D^{*}_{CNN}+w_2 D^{*}_{Trans}$, where $w_1$ and $w_2$ are the learned weights.

\subsection{Network optimization}
\label{Network optimization}

We leverage a counting loss and two novel quantity-oriented losses, \ie vision-text quantity alignment loss and cross-stream quantity ranking loss, for the network optimization.

\noindent \textbf{Counting loss.} We follow previous work~\citep{Xu_2023_CVPR} to leverage the counting loss $L_{count}$ to minimize the distance between the predicted density maps and the ground truth density map $D$. It is formulated as:
\begin{align}
\label{eqa:mse loss}
L_{count} &= \left \| D^{*}- D \right \|^{2}_2 + \left \| D^{*}_{CNN}- D \right \|^{2}_2 + \left \| D^{*}_{Trans}- D \right \|^{2}_2 
\end{align}

\noindent \textbf{Vision-Text quantity alignment loss.} 
This loss aims to align quantity-related vision and language information in the embedding space. This alignment is achieved by enforcing the similarities between visual and textual embeddings from the constructed vision-text pairs (Sec.~\ref{Prompts generate}) to conform to specific order.
Specifically, the positive vision-text pair, where the text prompt specifies the correct object count $a^p$, should exhibit the highest vision-text similarity score; in contrast, negative pairs should exhibit lower similarity scores than that of the positive pair. Moreover, among negative pairs, those with counterfactual counts $a^n$ closer to $a^p$ are expected to yield relatively higher similarity scores.
We suppose that $s^p$ denotes the cosine similarity between visual and textual embedding of the positive vision-text pair; $\left \{ {s^{n }_{i}|i \in [1,N]}\right \} $ represents the vision-text similarity scores of the negative pairs. $\left \{s_i^{n }\right \}$  can be split into two parts, the former half corresponds to the counterfactual counts $\left \{ {(a^p - k \Delta)|k \in [1,\frac{N}{2}]}\right \} $ while the latter half corresponds to $\left \{ {(a^p + k \Delta)|k \in [1,\frac{N}{2}]}\right \} $, respectively. Within each half, $s_i^{n}$ is ordered according to the descending order of their corresponding $k \Delta$ (the difference between the designated counterfactual count and the ground truth count). 
We can define the quantity alignment loss as:

\begin{small}
\begin{align}
\label{eqa:align loss}
L_{align} =\frac{1}{N}\sum_{i=1}^{N}&f(s^{n}_{i}-s^p)+ \nonumber \\
\frac{1}{N-2}\sum_{i=1}^{\frac{N}{2}-1}(f(s^{n}_{i}-s^{n}_{i+1})&+f(s^{n}_{\frac{N}{2}+i}-s^{n}_{\frac{N}{2}+i+1}))
\end{align}
\end{small}
where $f(\cdot)$ represents the $ReLU$ function. 
{The first term is similar to the conventional constraint in contrastive learning that forces the positive vision-text pair to have higher similarity than that of negative pairs; the loss value occurs when $s^{n}_{i} > s^p$. The second term constrains the similarities obtained from negative vision-text pairs to conform to specific order.  
It is designed to help the model capture fine-grained differences among numbers in images. As illustrated in Fig.\ref{fig:overview}(a), consider an image with one factual prompt ``a photo of 14 kiwis” and four counterfactual prompts, \ie, “a photo of 10/12/16/18 kiwis.” The corresponding vision-text similarity scores are expected to exhibit a quantity-aware ranking among negative pairs, such that the similarity between the image and “a photo of 12 kiwis” is higher than that between the image and “a photo of 10 kiwis” ; and the similarity for “16 kiwis” is higher than that for “18 kiwis”, as 12 and 16 are numerically closer to the ground-truth 14.} This loss helps enhance the quantity-awareness in the vision encoder (see also Fig.~\ref{fig:sim changes}). 

\noindent\textbf{Cross-stream quantity ranking loss.} Given the density maps $D^*_{CNN}$ and $D^*_{Trans}$ predicted from the two streams of DAC-decoder, we introduce a novel patch-level quantity ranking constraint to reduce their counting errors and increase their consistency. Specifically, we uniformly partition $D^*_{CNN}$ into $n$ non-overlapping local patches, which are ranked in descending order according to their ground truth object counts. Following this ranking order, we find the corresponding predicted object count for each patch of $D^*_{CNN}$ and write them as $V^{CNN}=\{ v^{CNN}_1,v^{CNN}_2,\cdots,v^{CNN}_n\}$. Similarly, the count list $V^{Trans}=\{ v^{Trans}_{1},v^{Trans}_{2},\cdots,v^{Trans}_{n}\}$ can be obtained from $D^*_{Trans}$ in the same manner. Our idea is that, for any two values $v_m$ and $v_{m+1}$, $v_m$ with a higher rank should ideally be greater than or equal to $v_{m+1}$, as this complies with their relative order indicated by their ground truth object counts. The same holds for $v^{Trans}$ and $v^{CNN}$. {To reinforce these orderings, we define a loss function to enforce the ranking constraint within each stream; more importantly, we introduce the cross-stream constraint between $V^{Trans}$ and $V^{CNN}$ to further enhance counting consistency across the streams.} The formula of this loss is:
\begin{small}
\begin{align}
\label{eqa:consistency loss}
L_{rank}= \frac{1}{n-l}\sum_{m=1}^{n-l}(f(v^{CNN}_{m+l}-v^{Trans}_{m})+f(v^{Trans}_{m+l}-v^{CNN}_{m})\nonumber \\
+f(v^{Trans}_{m+l}-v^{Trans}_{m})+f(v^{CNN}_{m+l}-v^{CNN}_{m}))
\end{align}
\end{small}
\noindent where $f(\cdot )$ is again the \emph{ReLU} function while $l$ denotes a patch interval. We set $l=5$ to allow some space between ranked items in the equation.  
If the space is too tight, \ie a small error in the prediction can flip the order and  cause the inconsistency between the predicted order and ground truth order, making it too sensitive. 
The former two parts of $L_{rank}$ enhance cross-stream counting consistency, while the latter two parts enforce the ranking constraints within each stream. 

The overall loss function is the combination of the above three losses:
\begin{align}
\label{eqa4}
L&= L_{count}+ \mu (L_{align}+ L_{rank})
\end{align}

{\noindent\textbf{Network inference.} During inference, we do not know the object count in the test image, hence the proposed quantity-oriented text prompts are no longer used ( Fig.~\ref{fig:overview}a: the dashed lines are removed); instead, likewise in~\citep{AminiNaieni23,kang2024vlcounter}, the category-specific prompt, such as ``a photo of birds”, is directly fed into the network to obtain the text embedding $F^C$ and is then interacted with the image embedding $F^I$  for the subsequent object counting (see Sec.~\ref{Prompts generate}).}
\section{Experiments}
\subsection{Experiment details}
\label{Experiments details}
\noindent\textbf{Datasets.} The training set of this work is FSC-147~\citep{ranjan2021learning} dataset. FSC-147 consists of 6,135 images from 147 categories. To evaluate our method, besides the validation and test sets of FSC-147, we follow previous works \citep{jiang2023clip,kang2024vlcounter} to use CARPK~\citep{hsieh2017drone}, PUCPR+~\citep{hsieh2017drone}, and ShanghaiTech~\citep{zhang2016single} datasets. These datasets exhibit significant differences from FSC-147, providing a robust validation of the model’s generalization. 
Specifically, CARPK contains nearly 90,000 cars from 4 different parking lots collected by drones. PUCPR+ contains images of 16,456 cars. The ShanghaiTech dataset is divided into SHA with 482 images collected from the Internet and SHB with 716 images captured on the busy streets of Shanghai.

\noindent\textbf{Implementation details.} We use the pre-trained DINOv2~\citep{oquab2023dinov2} with a ViT-B/14 backbone as our vision encoder and the pre-trained BERT-base~\citep{devlin2018bert} (top 9 blocks) as the text encoder. 
During training, we follow CounTX~\citep{AminiNaieni23} to freeze the weights of our text encoder, {while additionally fine-tuning the last six layers of the vision encoder}. The value interval $\Delta$ for generating counterfactual text prompts (Sec.~\ref{Prompts generate}) is determined as this: we identify the bin in which the ground truth count falls within $\{[0,10),[10,20),[20,50),[50,100),[100,200),[200,500),$ $[500,1000),[1000,\infty]\}$, then select the corresponding $\Delta$ from $\left \{ 1, 2, 3, 5,10,20 ,35,50\right \}$. 
We generate $N=7$ counterfactual text prompts to obtain negative vision-text pairs. Similar to previous works~\citep{jiang2023clip,peng2024single},  the image size ($H\times W$)  is set to 384 $\times$ 384 while the number of partitioned patches $n$ is 576. 
The loss weight $\mu$ in Eqn.~\ref{eqa4} is set to 0.1. 
Our model is trained using the AdamW \citep{loshchilov2017decoupled} optimizer with a learning rate of $1\times 10^{-4}$ and a decay factor of $\frac{1}{3}$. The training is performed over 200 epochs on a NVIDIA A40 GPU with a batch size of 32. Parameters are selected from validation. 

\noindent\textbf{Baseline.} To validate the effectiveness of our QUANet, we devise a Baseline. It generates the category prompt and uses the same encoders as QUANet to extract visual and textual embeddings, 
a CNN decoder is utilized to predict the density map. It is trained solely using the counting loss.

\noindent\textbf{Evaluation metrics.}
We follow previous works to use Mean Absolute Error (MAE) and Root Mean Square Error (RMSE) to evaluate the counting performance. 

\begin{table}[t]
\small
\caption{Quantitative comparisons on FSC-147.}
\label{tab:sota}
\setlength{\tabcolsep}{0.7mm}
\begin{tabular}{c|c|cc|cc}
\toprule
\multirow{2}*{Method} &\multirow{2}{*}{prompt}&\multicolumn{2}{c|}{VAL SET}& \multicolumn{2}{c}{TEST SET} \\ 
& &MAE$\downarrow$  &RMSE$\downarrow$  & MAE$\downarrow$  &RMSE$\downarrow$ \\ 
\midrule
GMN~\citep{lu2019class} &exemplar& 29.66& 89.91& 26.52& 124.57  \\
FamNet~\citep{ranjan2021learning}& exemplar& 23.75& 79.44& 24.90&112.68 \\
BMNet+~\citep{shi2022represent}& exemplar & 15.74& 58.53& 14.62& 91.83 \\
CounTR~\citep{liu2022countr}& exemplar&   13.13&49.83& 11.95& 91.23\\
LOCA~\citep{djukic2023low}& exemplar& 10.24&  32.56& 10.79& 56.97\\
 UPBC~\citep{lin2024fixed}& exemplar& 12.80& 48.65& 11.86&89.40\\
\midrule
ZSOC~\citep{Xu_2023_CVPR}& text& 26.93& 88.63&  22.09&115.17\\
CounTX~\citep{AminiNaieni23}& text&17.70& 63.61& 15.73& 106.88\\
CLIP-Count~\citep{jiang2023clip} & text  &  18.79& 61.18& 17.78&106.62\\
VLCounter~\citep{kang2024vlcounter} & text  &  18.06&65.13& 17.05&106.16\\
DAVE$_{prm}$~\citep{pelhan2024dave}& text&  15.48&  \textbf{52.57}& 14.90&103.42\\
VA-Count~\citep{zhu2024zero}& text& 17.87& 73.22& 17.88&129.31\\
CountDiff~\citep{hui2024class}& text& 15.50& 54.33& 14.83&103.15\\
 VLPG~\citep{10740342}& text& 16.05& 53.49& 17.60&97.66\\
\midrule
Baseline & text& 17.73& 65.56& 18.67&108.53\\
QUANet& text &  \textbf{14.22}&53.39& \textbf{13.24}&\textbf{97.24}\\
\bottomrule
\end{tabular}
\end{table}

\begin{figure*}
    \centering    
    \includegraphics[width=0.8\textwidth]{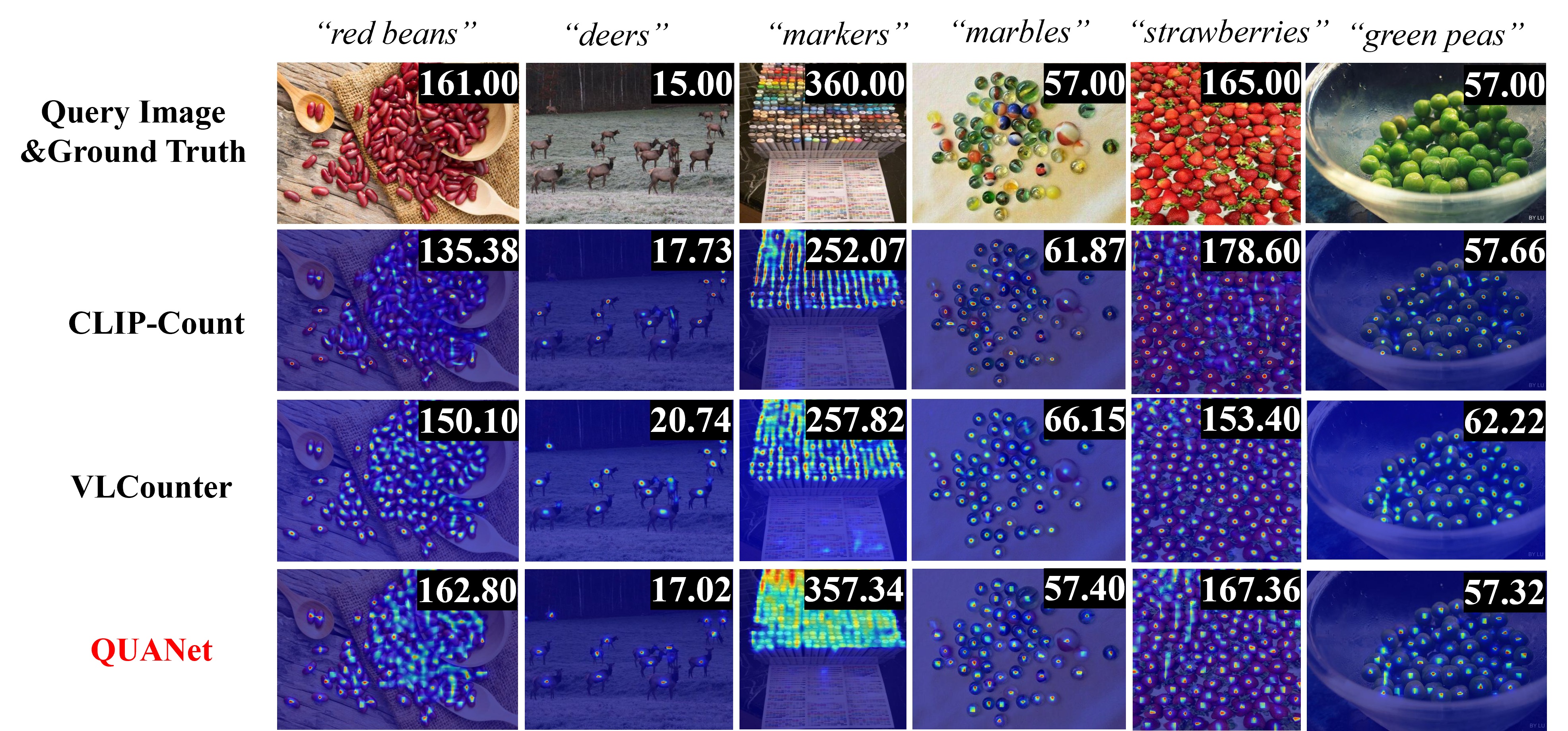}
    \caption{Visual comparison between different text-promptable object counting methods.  
    Counting values are shown at the right top of the query image and the predicted density map.}
    \label{fig:Visualization results}
\end{figure*}

\subsection{Comparison with state-of-the-art methods}
\label{Comparison with SOTA methods}
As shown in Tab.~\ref{tab:sota}, we compare the proposed method with six exemplar-based object counting methods (GMN~\citep{lu2019class}, FamNet~\citep{ranjan2021learning}, 
BMNet+~\citep{shi2022represent}, LOCA~\citep{djukic2023low},
CounTR~\citep{liu2022countr}, UPBC~\citep{liu2023point}) 
and eight text-promptable object counting methods (ZSOC~\citep{Xu_2023_CVPR}, CounTX~\citep{AminiNaieni23}, CLIP-Count~\citep{jiang2023clip}, VLCounter~\citep{kang2024vlcounter}, VA-Count~\citep{zhu2024zero}, DAVE$_{prm}$~\citep{pelhan2024dave}, CountDiff~\citep{hui2024class},VLPG~\citep{10740342}) on the FSC-147 dataset. First, we can observe that state-of-the-art exemplar-based  methods produce better results than text-promptable ones. 
Nevertheless, they require manually select exemplars, which is inconvenient. 
The performance gap between QUANet and exemplar-based counting methods is narrow. For instance, it is only inferior to two advanced models CounTR~\citep{liu2022countr} and LOCA~\citep{djukic2023low}. Second, QUANet achieves the best performance among text-promptable methods: on test set, it significantly decreases MAE by -1.59 and RMSE by -5.91 from very recent method CountDiff~\citep{hui2024class} and decreases MAE by -5.43 and RMSE by -11.29 from our Baseline. Qualitative results in Fig.\ref{fig:Visualization results} illustrate that QUANet produces high-quality density predictions across diverse scene distributions.

Next, we follow previous methods ~\citep{jiang2023clip, kang2024vlcounter,shi2022represent} to conduct cross-dataset validation by training QUANet on the FSC-147 dataset and evaluating it on the CARPK, PUCPR+ and ShanghaiTech datasets without any fine-tuning. As shown in Tab.~\ref{tab:car}, QUANet achieves the best results. 
Tab.~\ref{tab:the people} shows the results on the ShanghaiTech dataset. QUANet significantly decreases RMSE from 284.6 to 239.6 on SHA  and reduces MAE from 42.4 to 34.9 on SHB compared to VLPG~\citep{10740342}. These results demonstrate the strong generalizability of our method in the unknown scenario.

\begin{table}[!t]
\small
\centering
\caption{Cross-dataset evaluation on CARPK and PUCPR+.}
\label{tab:car}
\setlength{\tabcolsep}{0.7mm}
\begin{tabular}{c|c|cc|cc}
\toprule
\multirow{2}*{Method} &\multirow{2}{*}{prompt}&\multicolumn{2}{c|}{CARPK}& \multicolumn{2}{c}{PUCPR+} \\ 
& &MAE$\downarrow$  &RMSE$\downarrow$  & MAE$\downarrow$  &RMSE$\downarrow$  \\ 
\midrule
FamNet~\citep{ranjan2021learning}& exemplar& 28.84& 44.47& 87.54&117.68 \\
BMNet+~\citep{shi2022represent}& exemplar & 10.44& 13.77& 62.42& 81.74 \\
CounTR~\citep{liu2022countr}& exemplar&   5.75&7.45&  -&  - \\
\midrule
CLIP-Count~\citep{jiang2023clip} & text  &  11.96& 16.61& -&- \\
 CounTX~\citep{AminiNaieni23}& text& 11.64& 14.85& -&- \\
VLCounter~\citep{kang2024vlcounter} & text  &  6.46&8.68& 48.94&\textbf{69.08} \\
 VA-Count~\citep{zhu2024zero}& text& 10.63& 13.20& -&- \\
 CountDiff~\citep{hui2024class}& text& 10.32& 12.92& -&- \\
 VLPG~\citep{10740342}& text& 10.14& 13.79& -&-\\
\midrule
Baseline& text & 9.15& 11.66& 61.23& 87.35\\
QUANet& text &  \textbf{6.32}&\textbf{8.08}& \textbf{44.34}&70.25\\
\bottomrule
\end{tabular}
\end{table}

\begin{table}[!t]
\small
\centering
\caption{Cross-dataset evaluation on ShanghaiTech.}
\label{tab:the people}
\setlength{\tabcolsep}{0.7mm}
\begin{tabular}{c|cc|cc}
\toprule
\multirow{2}*{Method} &\multicolumn{2}{c|}{SHA}& \multicolumn{2}{c}{SHB} \\ 
& MAE$\downarrow$  &RMSE$\downarrow$  & MAE$\downarrow$  &RMSE$\downarrow$ \\
\midrule
RCC~\citep{hobley2022-LTCA} & 240.1& 366.9&  66.6&104.8\\
CLIP-Count~\citep{jiang2023clip} & 192.6&308.4& 45.7&77.4\\
VLPG~\citep{10740342}& 178.9& 284.6& 42.4&71.6\\
\midrule
Baseline& 194.1&302.8& 48.7&78.1\\
QUANet &  \textbf{140.2}&\textbf{239.6}& \textbf{34.9}&\textbf{59.9}\\
\bottomrule
\end{tabular}
\end{table}

\subsection{Ablation study}
\label{Ablation studies}
We follow previous methods~\citep{Xu_2023_CVPR,jiang2023clip,shi2022represent} to conduct ablation study on the validation set and test set of FSC-147 dataset. 

\noindent\textbf{Quantity-oriented text prompts} 

\noindent\emph{Effectiveness of quantity-oriented text prompts.} In our method, we introduce quantity-oriented text prompts (QTPs) to enhance the vision encoder's quantity awareness for object counting. In Tab.~\ref{tab:quantity text prompt}, we denote QUANet w/o QTPs as a variant where the QTPs are removed from the framework. The vision-text quantity alignment loss is also removed due to the absence of QTPs. The results show a significant increase in MAE, rising by +1.90 on the validation set and +1.86 on the test set, indicating the effectiveness of QTPs in enhancing the model's quantity awareness. 

\noindent\emph{Effectiveness of factual prompts in QTPs.} Next, we remove the factual prompts (FP) from QTPs to verify their contributions to the overall performance. This variant is denoted as QUANet w/o FP in Tab.~\ref{tab:quantity text prompt}. It increases MAE by +1.22 on the validation set and +1.65 on the test set, demonstrating its effectiveness. 

\noindent\emph{Design of counterfactual prompts in QTPs.} We further verify the design of counterfactual prompts in QTPs. As mentioned in Sec.~\ref{Experiments details}, we determine their incorrect counts based on a dynamic value interval $\Delta$ proportional to the ground truth count, ensuring the model to accurately distinguish subtle differences in counts.
In Tab.~\ref{tab:quantity text prompt} we use QUANet($\Delta=3/5/7/9)$ to denote a variant in which these incorrect counts are created by a fixed value interval. We can see these variants exhibit inferior performance compared to the original QUANet.
{ This is due to the highly varied distribution of object counts in the dataset (from 7 to 1912 in FSC-147). Using fixed $\Delta$ may lead to overfitting on certain samples while being ineffective on other samples. Therefore, we adopt an adaptive strategy for $\Delta$. Empirically, it is set in proportion to the lower bound of each count interval, which varies such as [10,20)...$[1000,\infty)$. We also offer two variants when the adaptive capacity is doubled (2$\Delta$) or reduced to half (0.5$\Delta$) in Tab.~\ref{tab:quantity text prompt}. Both perform only slightly inferior to our default setting, attesting the robust choice of our $\Delta$.}

\begin{figure*}[t]
\centering
\begin{tabular}{ccc}
\includegraphics[width=1.5in]{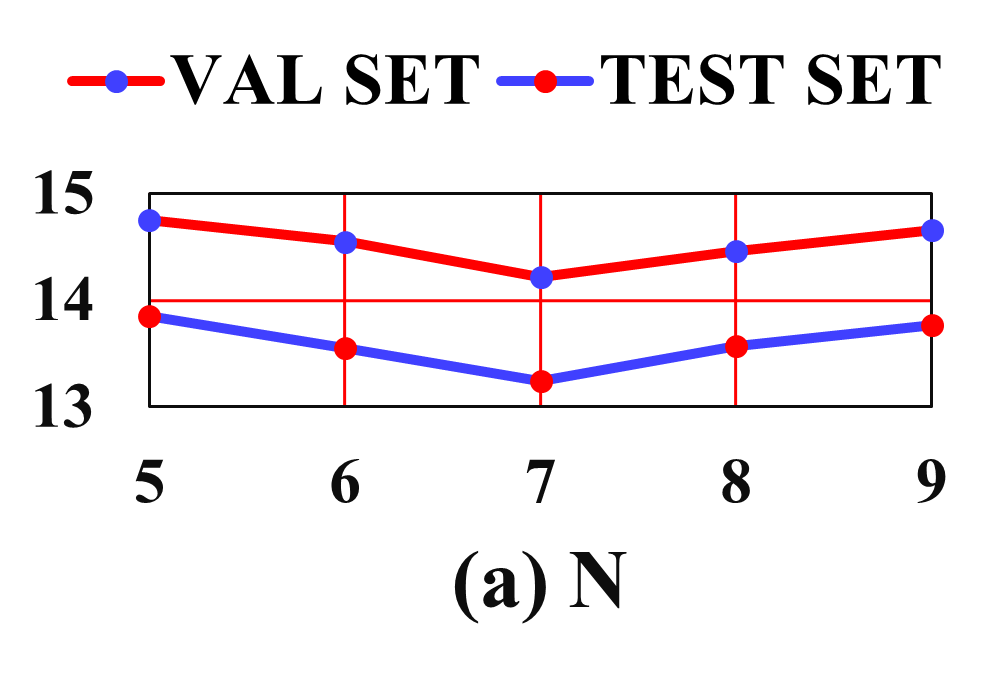} &
\includegraphics[width=1.5in]{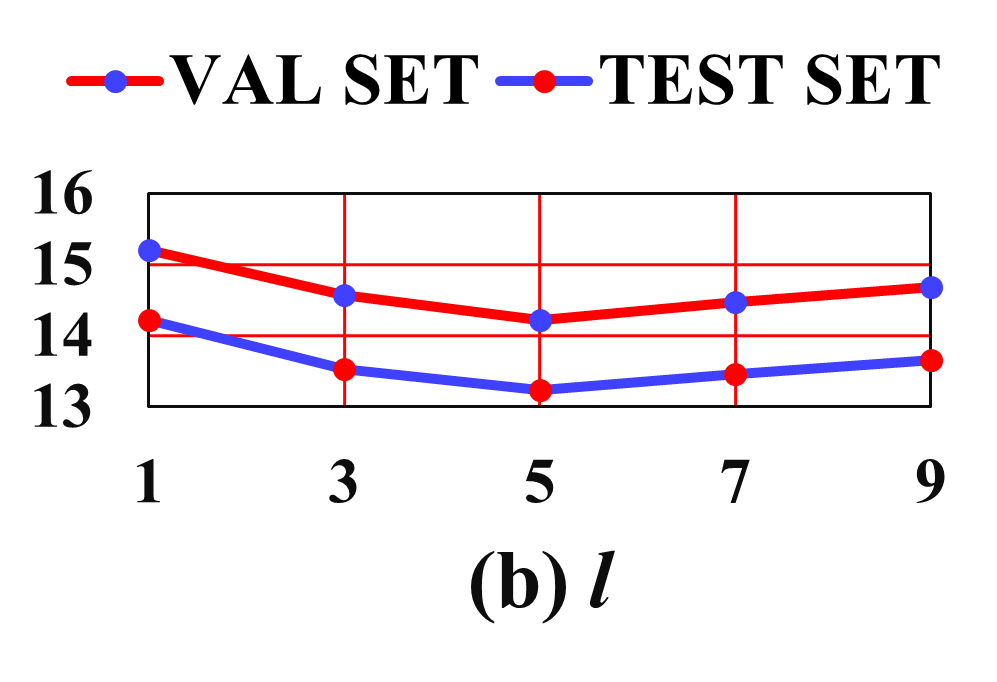} &
\includegraphics[width=1.5in]{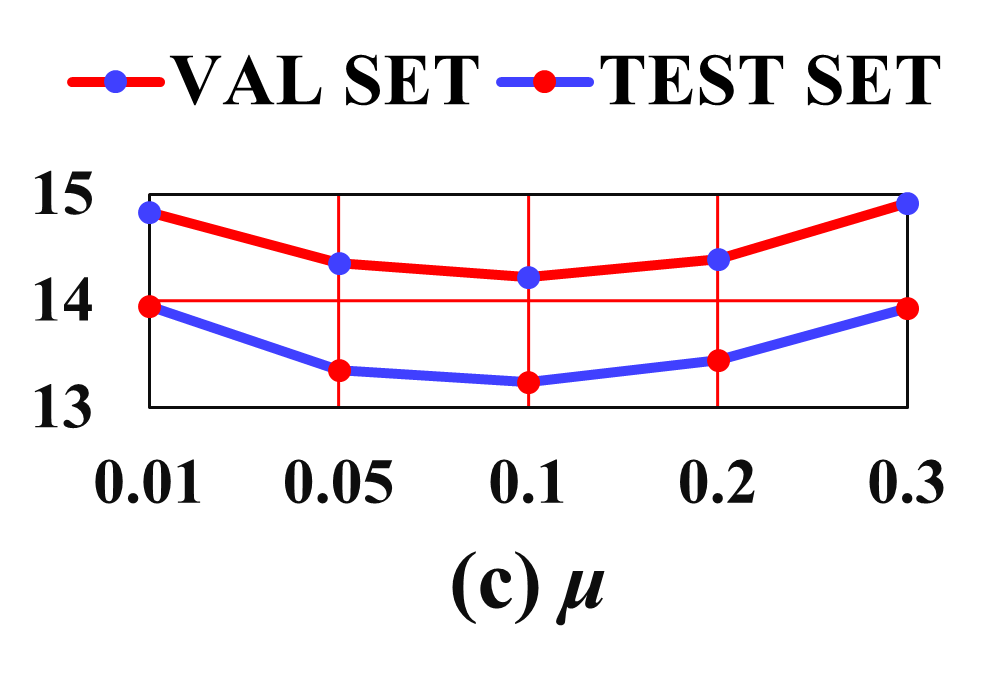} \\
\end{tabular}
\vspace{-10pt}
\caption{Parameter variation analysis on (a) the number $N$ of counterfactual text prompts, (b) the patch interval $l$ in the cross-stream quantity ranking loss and (c) $\mu$  in Eqn.~\ref{eqa4}. We present the MAE on FSC-147.}
\label{fig:mse}
\end{figure*}

Last, we vary the number of generated counterfactual QTPs, $N$,  from 5 to 9. As shown in Fig.~\ref{fig:mse}(a), we select the default $N=7$ as it performs the best on the validation set, and  concurrently also the best in testing. Notice if we remove all counterfactual text prompts, the alignment loss no longer exists, which is indeed QUANet w/o QTPs.

\noindent\emph{Quantity-oriented prompts \vs category-oriented prompts.} We verify the effectiveness of the proposed quantity-oriented prompts versus the category-oriented prompts. QUANet-CTP in Tab.~\ref{tab:quantity text prompt} denotes a variant of QUANet in which our quantity-oriented text prompts are replaced with the category-oriented prompts. Specifically, the factual prompt specifies the target object category in the query image, while the counterfactual prompts are created by randomly changing the object category names in the factual prompt.
We can see this variant also improves performance (\vs QUANet w/o QTPs) but is not comparable to our default QUANet.  Moreover, when combining both two types of prompts (denoted by QUANet-CQTP in Tab.~\ref{tab:quantity text prompt}), no cumulative benefits can be obtained.

\begin{table}[!t]
\small
\centering
\caption{Ablation study on the quantity-oriented text prompts.}
\label{tab:quantity text prompt}
\setlength{\tabcolsep}{0.7mm}
\begin{tabular}{c|cc|cc}
\toprule
  \multirow{2}*{Method}&\multicolumn{2}{c|}{VAL SET}& \multicolumn{2}{c}{TEST SET} \\ 
  &MAE$\downarrow$  &RMSE$\downarrow$  & MAE$\downarrow$  &RMSE$\downarrow$ \\ 
\midrule
QUANet w/o QTPs& 16.12& 61.81& 15.10& 103.16\\
QUANet w/o FP& 15.44& 59.32& 14.89&101.42\\
QUANet-CTP & 15.37& 59.32& 14.65&101.83\\
QUANet-CQTP & 14.89& 56.93& 13.93&99.39\\
\midrule
QUANet($\Delta=3$)& 15.91& 60.96& 14.92&100.83\\
QUANet($\Delta=5$)& 15.82& 60.91& 14.90&100.42\\
QUANet($\Delta=7$)& 15.13& 58.24& 14.20&99.17\\
QUANet($\Delta=9$)& 15.52& 59.47& 14.55 &100.56\\
QUANet(2$\Delta$)&14.81 &57.46 & 13.77&98.48\\
QUANet($0.5\Delta$)&14.59 &56.01 & 13.63&98.26\\
\midrule
QUANet& \textbf{14.22}& \textbf{53.39}& \textbf{13.24}&\textbf{97.24}\\
\bottomrule
\end{tabular}
\end{table}

\noindent\textbf{DAC-decoder} 

\noindent\emph{Effectiveness of DAC-decoder.} We first investigate the effectiveness of the dual-stream structure of the proposed DAC-decoder. In Tab.~\ref{tab:qca-decoder}, we denote the DAC-decoder without the CNN stream as DAC-decoder w/o CNN and the variant without the Transformer stream as DAC-Decoder w/o Trans. Both variants show a clear performance decline.

We further investigate the complementary nature  between Transformer decoder and CNN decoder through 
the multi-level metric GAME~\citep{li2018csrnet}. GAME is defined {as $\text{GAME}(L)= \frac{1}{n} \sum_{i=1}^{n}(\sum_{l=1}^{4^L}|y_{l}^{*}-y_{l}^{gt}|)$, where $n$ is the number of test images; $y_{l}^{*}$ and $y_{l}^{gt}$ denote the predicted and ground truth object counts for the $l$-th region in $i$-th image, respectively. The image is evenly divided into $4^L$ non-overlapping grids, with larger $L$ indicating finer regional evaluation. GAME measures both the global and local counting errors. }
Tab.\ref{tab:GAME Metric} presents the respective results of single-steam decoders: the Transformer decoder, DAC-decoder w/o CNN
excels in capturing global information (G0 and G1); the CNN decoder, DAC-Decoder w/o Trans, is more effective at estimating local counts in smaller patches (G2 and G3). This observation leads to the design of the T2C-adapters and gating network to transfer and combine the strength of both decoders for final prediction, which scores the lowest across G0$\sim$G3.

Last, it should be noted that our DAC-decoder is not a simple ensemble of more models and parameters. As being said, the knowledge has to be carefully transferred between the Transformer stream and CNN stream in order to successfully cast their complementary nature into realistic benefit. 
If we simply average two CNN streams, denoted as QUANet(DC) in Tab.~\ref{tab:qca-decoder}, (or two Transformers, denoted as QUANet(DT)), the benefit over single-stream variant of QUANet is insignificant.

\begin{table}[!t]
\small
\centering
\caption{Ablation study on the DAC-decoder.}
\label{tab:qca-decoder}
\setlength{\tabcolsep}{0.7mm}
\begin{tabular}{c|cc|cc}
\toprule
\multirow{2}*{Method}&\multicolumn{2}{c|}{VAL SET}& \multicolumn{2}{c}{TEST SET} \\ 
&MAE$\downarrow$  &RMSE$\downarrow$  & MAE$\downarrow$  &RMSE$\downarrow$ \\
\midrule
DAC-decoder w/o {Trans}& 16.84& 61.57& 16.92& 112.88\\
DAC-decoder w/o {CNN}& 16.05& 62.71& 16.30&106.17\\
 QUANet(DC)& 15.79& 59.42& 15.30&107.98\\
 QUANet(DT)& 15.36& 61.33& 15.45&103.47\\ 
\midrule
DAC-decoder w/o {T2C}& 15.12& 55.46& 14.73& 98.63\\
DAC-decoder w/ {C2T}& 15.92& 55.32& 15.42&99.55\\
DAC-decoder w/ {BiD}& 15.91& 58.93& 14.00&104.81\\
DAC-decoder w/o CE& 14.91& 57.01& 14.39&100.93\\
DAC-decoder w/o CA& 14.83& 55.81& 14.26&99.14\\
\midrule
DAC-decoder w/ avg-$w$& 15.31& 60.09& 13.86&101.07\\
\midrule 
DAC-decoder (QUANet)&   \textbf{14.22}&\textbf{53.39}& \textbf{13.24}& \textbf{97.24}\\
\bottomrule
\end{tabular}
\end{table}

\noindent \emph{Effectiveness of T2C-adapter.} We further investigate the effectiveness of the proposed T2C-adapter. This adapter transfers knowledge from the Transformer stream into the CNN stream on the patch level,
providing contextual information for the overall object count. As shown in Tab.~\ref{tab:qca-decoder}, removing the T2C-adapters (denoted as DAC-decoder w/o T2C) results in degraded performance across both datasets. Moreover, if we reverse the knowledge transformation direction (denoted as DAC-decoder w/ {C2T} or apply bidirectional knowledge transformation (denoted as DAC-decoder w/ {BiD}), the MAE substantially increases on the test set. 
Furthermore, we present the result of T2C-adapter without the channel excitation/cross attention block, denoted by DAC-decoder w/o CE/CA in Tab.~\ref{tab:qca-decoder}. Removing either block leads to a performance drop, yet the performance still outperforms the variant DAC-decoder w/o T2C, indicating that both contributes to the overall performance.

\noindent \emph{Effectiveness of Gating net.} The gating net assigns weights to aggregate the outputs of the CNN and Transformer streams in the DAC-decoder. If we remove it but take the average of the outputs as the final prediction, the MAE and RMSE will be significantly increased (see DAC-decoder w/ avg-$w$ in Tab.~\ref{tab:qca-decoder}), indicating the effectiveness of our gating net.

\begin{table}[!t]
\small
\centering
\caption{Ablation study of the GAME results on the validation set of FSC-147.}
\label{tab:GAME Metric}
\setlength{\tabcolsep}{0.7mm}
\begin{tabular}{c|c|c|c|c} 
\toprule
Method&G0$\downarrow$&G1$\downarrow$& G2$\downarrow$&G3$\downarrow$\\ 
\midrule
DAC-decoder w/o {Trans}& 16.84& 18.91& 21.42& 26.96\\
DAC-decoder w/o {CNN}& 16.05& 18.17& 21.94&28.56\\
DAC-decoder (QUANet)& 14.22& 16.35& 19.46&25.11\\
\bottomrule
\end{tabular}
\end{table}

\begin{table}[!t]
\small
\centering
\caption{Ablation study on the generalizability of DAC-decoder.}
\label{tab:dac generalizability}
\setlength{\tabcolsep}{0.7mm}
\begin{tabular}{c|cc|cc}
\toprule
\multirow{2}*{Method}&\multicolumn{2}{c|}{VAL SET}& \multicolumn{2}{c}{TEST SET} \\ 
&MAE$\downarrow$  &RMSE$\downarrow$  & MAE$\downarrow$  &RMSE$\downarrow$ \\
 \midrule
 CLIP-Count~\citep{jiang2023clip} & 18.79& 61.18& 17.78&106.62\\
 VLCounter~\citep{kang2024vlcounter} & 18.06& 65.13& 17.05&106.16\\
 \midrule
 CLIP-Count w/ DAC & 17.17& 59.62& 16.20 &102.23\\
 VLCounter w/ DAC & 16.51& 62.40& 15.56&101.67\\
 \bottomrule
\end{tabular}
\end{table}

\noindent {\emph{Generalizability of DAC-decoder.}We replace the original decoders of CLIP-Count~\citep{kang2024vlcounter} and VLCounter~\citep{kang2024vlcounter} with our DAC-decoder in Tab.\ref{tab:dac generalizability}. Both outperform their original versions, showcasing the generalizability of DAC-decoder. }

\noindent\textbf{Loss functions} 

\noindent \emph{Vision-text quantity alignment loss $L_{align}$.} 
To understand the contribution of each term in the alignment loss $L_{align}$, we present two variants by removing the first or second term from Eqn.\ref{eqa:align loss} to investigate the effectiveness of both, namely QUANet($L_{align}$ w/o FT/ST) in Tab.~\ref{tab:quantity consistency loss}. Both leads to performance drop, attesting their effectiveness.
{The first term $\sum_{i=1}^{N}f(s^{n}_{i}-s^p)$ essentially plays the same role to a standard vision–text contrastive loss as in $L_{vtc}$~\citep{radford2021learning, paiss2023teaching}, which treats all negative samples equally. We found their effects to be empirically similar: using only the first term (QUANet($L_{align}$ w/o ST)) yields similar (slightly better) results to replacing $L_{align}$ with $L_{vtc}$ (QUANet($L_{align} \rightarrow L_{vtc}$)). 
Next, the second term enforces a ranking among the similarity scores of those negative vision-text pairs. This quantity-aware refinement offers clear improvement in the overall performance. 
}

\noindent \emph{Cross-stream quantity ranking loss $L_{rank}$.} This loss is proposed to optimize patch-level quantity ranks both within and across the predictions of the two streams in the DAC-decoder. As shown in Tab.~\ref{tab:quantity consistency loss}, omitting $L_{rank}$ leads to deteriorated results on both datasets, \eg, the MAE increases by +1.70 on the validation set and +1.81 on the test set. This demonstrates the superiority of $L_{rank}$ in reducing counting errors.
Furthermore, we introduce a variant of $L_{rank}$ in which the patch-level ranking constraints are applied solely within each independent stream of DAC-decoder. This variant, denoted by QUANet($L_{rank} \rightarrow L_{srank}$) in Tab.~\ref{tab:quantity consistency loss}, leads to a performance decline, with MAE increased by +1.12 on the test set. We also experiment with maintaining only the cross-stream ranking constraint, omitting the ranking constraint within each stream. This variant is denoted as QUANet($L_{rank} \rightarrow L_{crank}$) in Tab.~\ref{tab:quantity consistency loss}, and it is also inferior to the original $L_{rank}$.  Last, we vary the patch interval $l$ in $L_{rank}$ (Eqn.\ref{eqa:consistency loss}) from 1 to 9. As shown in Fig.~\ref{fig:mse}(b), QUANet performs the best when $l=5$, which is our default setting.

\noindent \emph{Loss weight between $L_{align}$ and $L_{rank}$.} We investigate the influence of the loss weight $\mu$ in Eqn.~\ref{eqa4}. As shown in Fig.~\ref{fig:mse}(c), our QUANet performs the best when $\mu=0.1$ on the validation set (default), and concurrently also the best on the test set.

\begin{table}[!t]
\small
\centering
\captionsetup{width=\textwidth}
\caption{Ablation study on the loss functions.}
\label{tab:quantity consistency loss}
\setlength{\tabcolsep}{0.7mm}
\begin{tabular}{c|cc|cc}
\toprule
 \multirow{2}*{Method}&\multicolumn{2}{c|}{VAL SET}& \multicolumn{2}{c}{TEST SET} \\ 
 &MAE$\downarrow$  &RMSE$\downarrow$  & MAE$\downarrow$  &RMSE$\downarrow$ \\
\midrule
 QUANet($L_{align}$ w/o FT)&15.29 &56.85 & 13.95&99.87\\
 QUANet($L_{align}$ w/o ST)&15.01 &58.02 & 13.81&99.44\\
 QUANet($L_{align}\rightarrow L_{vtc}$)& 15.42& 60.27& 13.84& 102.21\\
 \midrule
QUANet w/o $L_{rank}$& 15.92& 55.77& 15.05&103.74\\
QUANet($L_{rank} \rightarrow L_{srank}$)& 15.27& 54.70& 14.36&98.22\\
QUANet($L_{rank} \rightarrow L_{crank}$)& 14.74& 56.20& 13.41&99.73\\
\midrule
QUANet& \textbf{14.22}& \textbf{53.39}& \textbf{13.24}&\textbf{97.24}\\
 \bottomrule
\end{tabular}
\end{table}

\subsection{Additional analysis}
\noindent \textbf{Complexity.} We compare the parameters and inference speed between QUANet and current methods ~\citep{kang2024vlcounter,AminiNaieni23,pelhan2024dave,zhu2024zero}. As shown in Tab.~\ref{tab:Params(MB)}, QUANet shows comparable complexity to others. 
Notice 1) DAVE$_{prm}$ and VA-Count are  multi-stage works, \eg VA-Count employs GroundingDINO, CLIP, \etc, we can only roughly estimate their complexities; 2) the DAC-decoder of QUANet account for less than 10\%  of the total parameters. 

\begin{table}[!t]
\small
\centering
\caption{Comparison on parameters and inference speed.}
\label{tab:Params(MB)}
\setlength{\tabcolsep}{0.7mm}
\begin{tabular}{c|cc} 
\toprule
 Method&Params(MB) $\downarrow$ &Speed(fps) $\uparrow$ \\
\midrule
VLCounter~\citep{kang2024vlcounter}& 577.4& 17.8\\
CounTX~\citep{AminiNaieni23}& 614.5& 22.5\\
 DAVE$_{prm}$~\citep{pelhan2024dave}& $\sim$750 &$\sim$ 6 \\
VA-Count~\citep{zhu2024zero}& $>$1000 & $<$ 6 \\
QUANet& 773.4& 22.8\\
\bottomrule
\end{tabular}
\end{table}

\noindent \textbf{Visualization}

\begin{figure}
    \centering
    \includegraphics[width=0.5\textwidth]{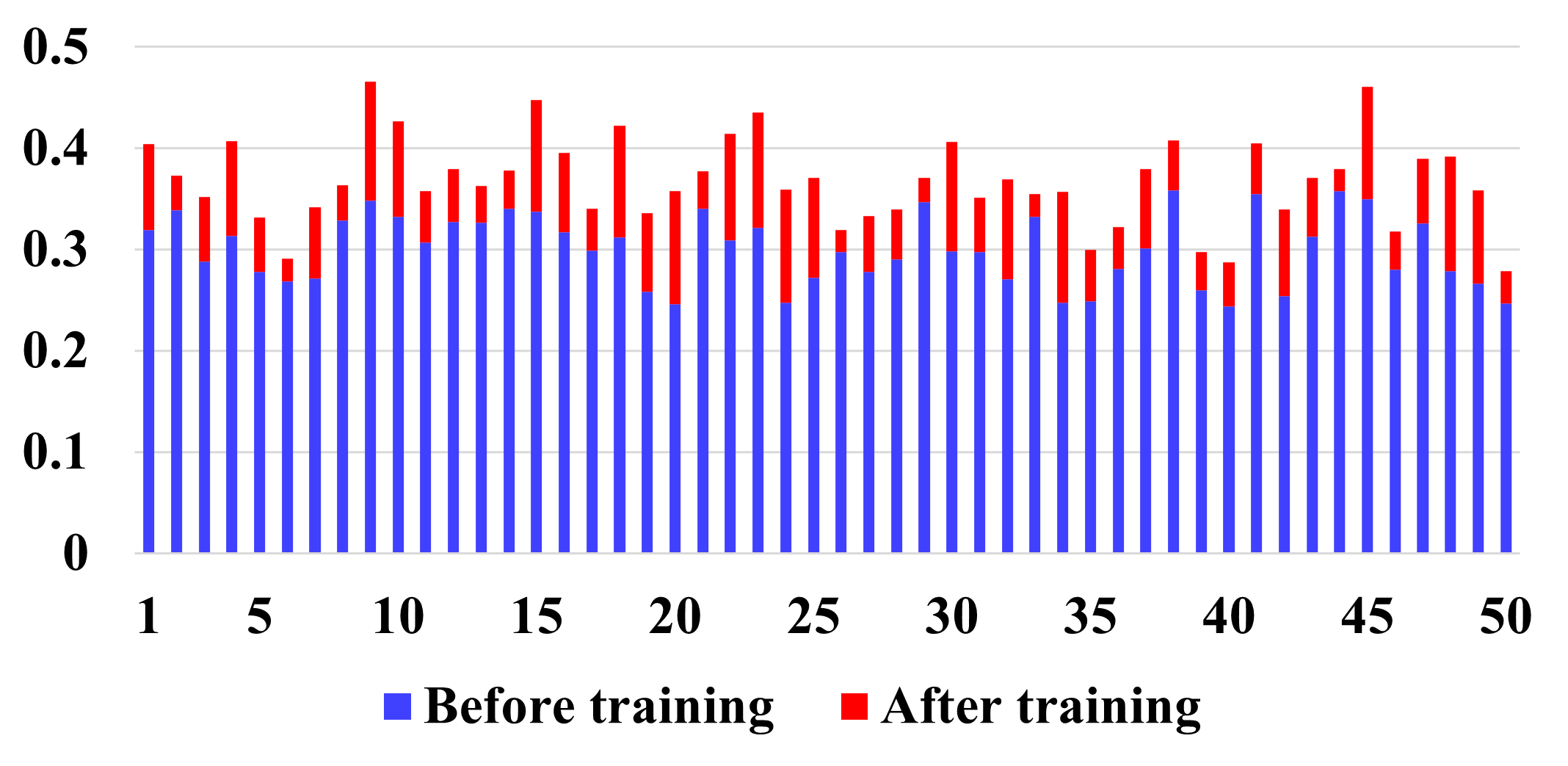}
    \caption{Similarities between positive vision-text pairs.}
    \label{fig:sim changes}
\end{figure}

\noindent \emph{Similarity scores for vision-text pairs.} 
To verify the enhancement of quantity-awareness of the vision encoder, we calculate the cosine similarity between embeddings of the factual text prompt and the corresponding image global feature. The results show that the average similarity over all vision-text pairs increases by 6\% after the model training. As illustrated in Fig.~\ref{fig:sim changes}, we select 50 samples to demonstrate the clear improvement, attesting the enhancement of quantity-awareness in vision encoder.

\begin{figure}[t]
  \centering
  \includegraphics[width=0.5\textwidth]{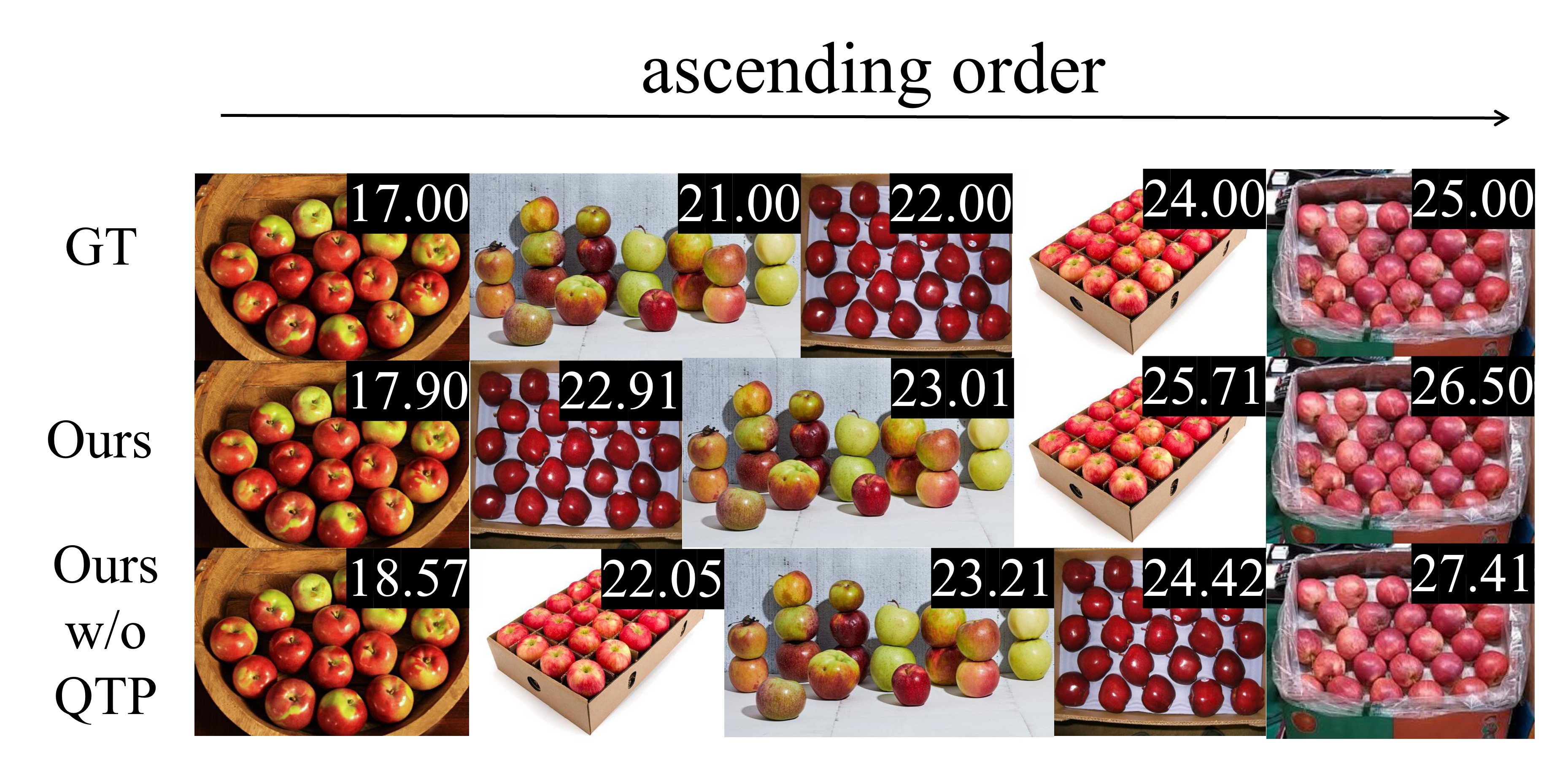}
   \caption{Sample sequences with ranked predictions.} 
   \label{fig:order}
\end{figure}

\noindent \emph{Preservation of numerical ranks.} 
We respectively sort images for each object class in ascending order based on their ground-truth and predicted counts, forming two ranked sequences per class.  
We then compute the mean Average Precision (mAP) \citep{superfeatures} between the two sequences across all classes. Compared to QUANet w/o QTPs, QUANet improves mAP from 80.10 to 85.70, indicating that it better comprehends numerical differences and thereby preserves numerical ranks among images. Visual examples are in Fig.~\ref{fig:order}.

\begin{figure}[t]
  \centering
  \includegraphics[width=0.5\textwidth]{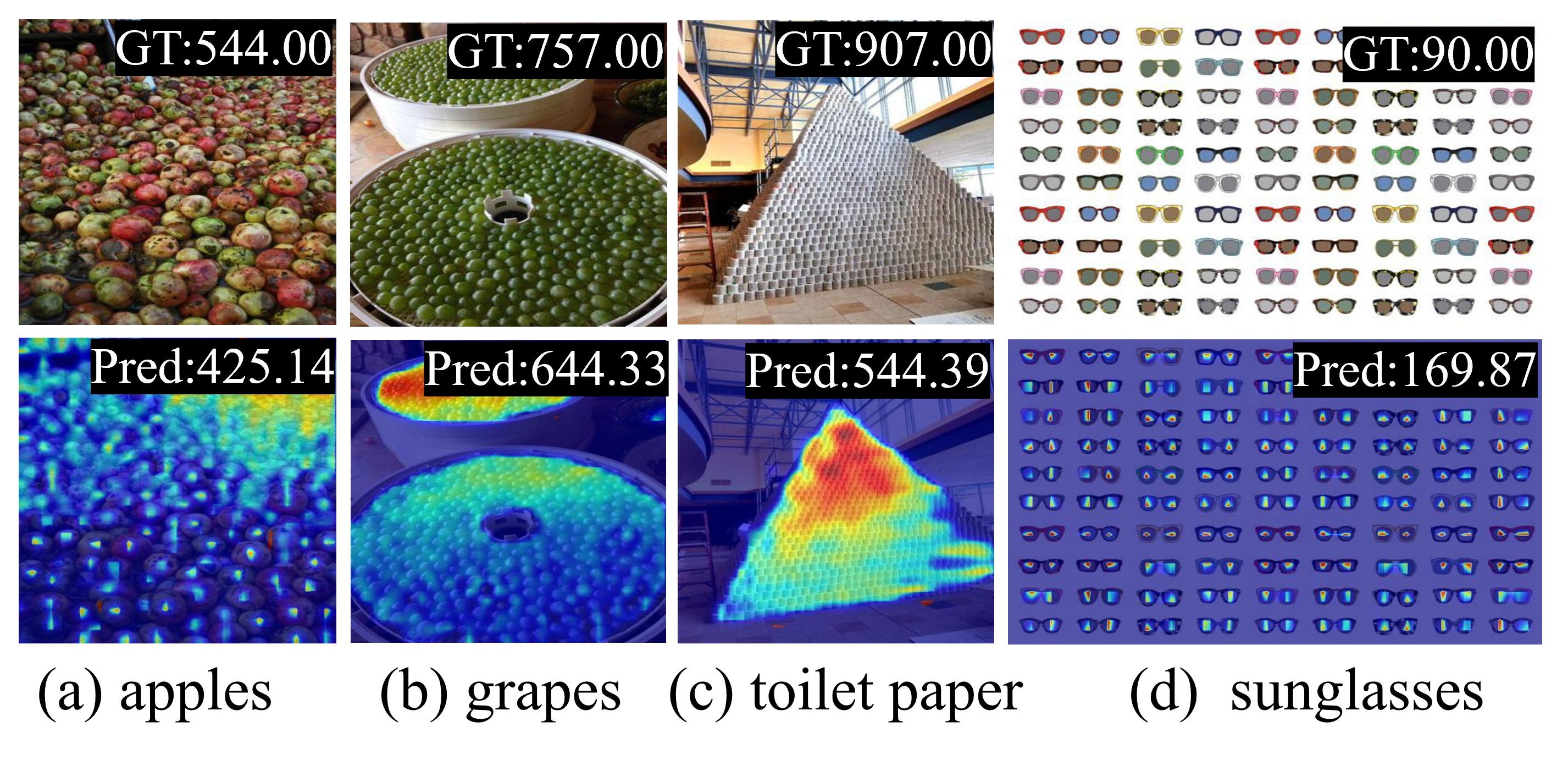}
   \caption{Failure cases of our QUANet.} 
   \label{fig:failure case}
\end{figure}

\noindent \emph{Failure case analysis.} Finally, we present some failure cases in Fig.\ref{fig:failure case} and summarize the key limitations. First, when objects are heavily stacked or severely occluded, { as shown in Fig.\ref{fig:failure case}(a), (b) and (c), } it can be challenging for QUANet to distinguish them. Second, for special cases such as spectacles, {as shown in Fig.\ref{fig:failure case}(d), }QUANet mistakenly counts each eyeglass as a separate object.

\section{Conclusion}
In this work, we propose a novel text-promptable object counting method that incorporates quantity-related descriptions to enhance the model's quantity awareness. Given a query image, our QUANet begins by generating quantity-oriented text prompts to describe the numbers of objects of the target category in the image. We construct a factual prompt with the correct object count to form positive vision-text pair and several counterfactual prompts with incorrect counts to create negative pairs. Next, we extract textual and visual embeddings from the prompts and the image using the VLM's encoders. A novel quantity alignment loss is applied to the embeddings to encourage vision-text alignment, thereby improving the vision encoder’s understanding of quantity-related visual features. To predict the density map, we introduce the DAC-Decoder, which encodes complementary global and local features through Transformer and CNN streams. Several layer-wise T2C-adapters are designed to facilitate the iterations of these features to improve the counting performance. Moreover, a cross-stream quantity ranking loss is employed to optimize patch-level quantity consistency between predictions. Experiments on four datasets show the effectiveness and superiority of our method over state of the art.

{
    \small
    \bibliographystyle{ieeenat_fullname}
    \bibliography{main}
}

\end{document}